\pdfoutput=1
\documentclass[10pt,conference]{IEEEtran}
\IEEEoverridecommandlockouts

\newif\ifanonymous
\anonymousfalse  % Camera-ready: de-anonymized

\usepackage{cite}
\usepackage{amsmath,amssymb,amsfonts}
\usepackage{algorithmic}
\usepackage{graphicx}
\usepackage{textcomp}
\usepackage{xcolor}

\usepackage{booktabs}
\usepackage{tabularx}
\usepackage{siunitx}
\usepackage{enumitem}
\usepackage{makecell}
\usepackage{multirow}
\usepackage{url}

\usepackage[colorlinks=true,breaklinks=true,
            linkcolor=black,citecolor=black,filecolor=black,menucolor=black,
            urlcolor=blue]{hyperref}
\hypersetup{
  pdftitle={Early-Stopping Thresholds for ES-HyperNEAT: A Data-Driven Approach from Fitness Dynamics},
  pdfauthor={Romain Claret, Arthur Gygax, Michael O'Neill, Paul Cotofrei, Pascal Felber},
  pdfsubject={Neuroevolution; ES-HyperNEAT; hyperparameter optimization; early stopping},
  pdfkeywords={Data-driven pruning, early stopping, ES-HyperNEAT, hyperparameter optimization, neuroevolution}
}

\def\BibTeX{{\rm B\kern-.05em{\sc i\kern-.025em b}\kern-.08em
    T\kern-.1667em\lower.7ex\hbox{E}\kern-.125emX}}

\makeatletter
\def\ps@IEEEtitlepagestyle{%
  \let\@oddhead\@empty
  \let\@evenhead\@empty
  \def\@oddfoot{\mycopyrightnotice}%
  \let\@evenfoot\@empty
}
\def\mycopyrightnotice{%
  \begin{minipage}{\textwidth}\footnotesize
    \copyright~2026 IEEE. Personal use of this material is permitted. Permission from
    IEEE must be obtained for all other uses, in any current or future media, including
    reprinting/republishing this material for advertising or promotional purposes,
    creating new collective works, for resale or redistribution to servers or lists, or
    reuse of any copyrighted component of this work in other works.
  \end{minipage}%
}
\makeatother

\begin{document}

\title{Early-Stopping Thresholds for ES-HyperNEAT: \\ A Data-Driven Approach from Fitness Dynamics}

\ifanonymous
  \author{Anonymous Author(s)}
\else
  \author{\IEEEauthorblockN{Romain Claret\IEEEauthorrefmark{1}\IEEEauthorrefmark{2},
    Arthur Gygax\IEEEauthorrefmark{1}\IEEEauthorrefmark{3},
    Michael O'Neill\IEEEauthorrefmark{4},
    Paul Cotofrei\IEEEauthorrefmark{1}\IEEEauthorrefmark{2},
    and Pascal Felber\IEEEauthorrefmark{1}\IEEEauthorrefmark{3}}
  \IEEEauthorblockA{\IEEEauthorrefmark{1}University of Neuch\^atel, Switzerland}
  \IEEEauthorblockA{\IEEEauthorrefmark{2}Information Management Institute}
  \IEEEauthorblockA{\IEEEauthorrefmark{3}Computer Science Department}
  \IEEEauthorblockA{\IEEEauthorrefmark{4}Natural Computing Research \& Applications Group,
    University College Dublin, Ireland}
  \thanks{Romain Claret and Arthur Gygax contributed equally to this work.}
  \thanks{Corresponding author: Romain Claret (romain.claret@unine.ch).}
  }
\fi

\maketitle

\begin{abstract}
Most hyperparameter configurations for Evolvable-Substrate HyperNEAT (ES-HyperNEAT) produce networks that stagnate at random-guessing performance, wasting computational resources. We frame early stopping as binary classification on early fitness trajectories: for each trial, we compute the cumulative median of best-per-generation fitness and test it against a threshold derived by maximizing the $F_1$ score on an initial 90-trial dataset. The resulting rule (generation $G^*{=}3$, threshold $T^*{=}0.140$) achieves $F_1{=}0.872$ on 180 independent validation trials, retaining over 90\% of successful trials while cutting computational cost by 41.6\%. Compared to Hyperband, our domain-specific rule is 64\% more efficient with higher mean fitness, though Hyperband occasionally discovers higher peak solutions. On a converged search population the rule becomes too aggressive (recall 31.1\%), motivating adaptive thresholds. The specific thresholds are ES-HyperNEAT-specific, but the methodology, deriving stopping criteria from fitness dynamics classification, is applicable to other evolutionary algorithms with stagnation-prone hyperparameter spaces.
\end{abstract}


\section{Introduction}

Neuroevolutionary algorithms, such as the NeuroEvolution of Augmenting Topologies (NEAT)~\cite{stanley2002evolving} and its indirect encoding extension ES-HyperNEAT~\cite{risi2012enhanced}, can evolve neural network architectures without backpropagation. Their performance, however, depends heavily on hyperparameter choices, and the irregular search space of ES-HyperNEAT makes selecting good configurations difficult.

The Tree-structured Parzen Estimator (TPE)~\cite{bergstra2011algorithms} offers a structured way to search such spaces. Previous work~\cite{claret2024tpe} showed that TPE outperforms random search on ES-HyperNEAT classification benchmarks including MNIST, reaching higher accuracy with fewer evaluations by learning the structure of the search space.

One problem persists regardless of the search method: most configurations produce trials that show no learning and stagnate near random-guessing performance. These dead-end runs waste time without informing the search.

We ask three questions. (1)~Can early fitness dynamics reliably predict whether an ES-HyperNEAT trial will succeed or fail? (2)~How early in the evolutionary run can this prediction be made? (3)~Does a domain-specific stopping rule derived from these dynamics outperform general-purpose pruning schedules?

We propose an early-stopping strategy that kills unpromising runs before they finish. We observe that failing trials show a characteristic flat fitness trajectory within the first few generations, and we frame the pruning decision as binary classification: given the median best fitness up to a cutoff generation, is this trial going to succeed or not? The resulting rule is simple and interpretable, and it keeps most successful trials while cutting wasted computation.

We validate the rule on held-out trials and show that it generalizes. Integrating early stopping with TPE-driven search speeds up exploration without degrading solution quality.

\section{Related Work}

The optimization of neural networks via evolutionary algorithms, a field known as neuroevolution, offers an alternative to gradient-based methods. The NeuroEvolution of Augmenting Topologies (NEAT) algorithm~\cite{stanley2002evolving} evolves both the weights and the structure of networks, starting from minimal configurations. This approach of complexification, where topologies grow to match the problem's complexity, helps to minimize the dimensionality of the search space. ES-HyperNEAT~\cite{risi2012enhanced} extends this paradigm with an indirect encoding that generates regular network patterns on an evolvable substrate, allowing it to handle high-dimensional problems. The performance of these algorithms, however, depends on many interacting hyperparameters, making manual tuning impractical. We use ES-HyperNEAT with TPE-based hyperparameter optimization.

For hyperparameter optimization (HPO), Bayesian Optimization (BO) is a sample-efficient strategy~\cite{shahriari2015taking}. Unlike grid or random search~\cite{bergstra2012random}, BO builds a probabilistic surrogate model of the objective function and uses it to select configurations likely to yield improvements. The Tree Structured Parzen Estimator (TPE)~\cite{bergstra2011algorithms}, a popular BO algorithm, models separate probability distributions $\ell(x) = p(x \mid y < y^*)$ for well-performing configurations and $g(x) = p(x \mid y \geq y^*)$ for the remainder, then maximizes the ratio $\ell(x)/g(x)$ to propose new candidates. Unlike GP-based BO, TPE handles categorical and conditional variables natively and scales to large trial budgets, making it well suited to the mixed hyperparameter space of ES-HyperNEAT. TPE outperforms random search and other BO methods on many ML tasks. These methods improve search efficiency but do not reduce the cost of individual evaluations.

The high cost of evaluating many configurations has led to the development of early-stopping, or pruning, strategies. These methods aim to terminate unpromising trials early, reallocating computational resources to more promising candidates. Throughout this paper, ``pruning'' refers to early termination of hyperparameter trials, not network pruning (removing connections from neural networks), which is a separate technique in neuroevolution. Prominent examples include the Successive Halving Algorithm (SHA)~\cite{jamieson2016non} and Hyperband~\cite{li2017hyperband}, which periodically evaluate a portfolio of configurations and discard a fixed proportion of the worst performers. Other approaches attempt to predict the final performance of a trial by extrapolating its initial learning curve~\cite{domhan2015speeding}, allowing for early termination if the predicted outcome is not competitive. Frameworks like Optuna have integrated such mechanisms, for instance, by stopping trials that perform worse than the median of their peers at an intermediate step~\cite{akiba2019optuna}.

These general-purpose pruning methods work well but rely on fixed-elimination schedules or complex predictive models. We take a different approach: instead of predefined schedules or learning-curve extrapolation, we derive a pruning rule directly from the fitness dynamics of ES-HyperNEAT. Framing the pruning decision as binary classification, we obtain an interpretable, empirically grounded rule that separates successful from unsuccessful trials at the earliest possible generation.

ES-HyperNEAT~\cite{risi2012enhanced} evolves both weights and topology using an indirect encoding: a Compositional Pattern-Producing Network (CPPN)~\cite{stanley2007compositional} generates connectivity patterns, and an adaptive substrate mechanism sets topology through quadtree variance analysis. Unlike fixed-topology networks where a few learning-rate and regularization parameters dominate, ES-HyperNEAT exposes 16 interacting hyperparameters spanning NEAT mutation rates, population size, substrate depth, and variance thresholds for quadtree subdivision. Most configurations in this space produce networks that never learn, making ES-HyperNEAT a challenging and representative testbed for early-stopping research.

\section{Experimental Setup}

Given the stagnation-prone nature of the ES-HyperNEAT hyperparameter space described above, we introduce an early-stopping strategy to terminate unpromising trials before completion. The pruning decision is formulated as a threshold test on the cumulative median best fitness (Equation~\ref{eq:prune_rule}). The methodology has two phases: Phase~1 (\textit{Derivation}) establishes the pruning criterion via classification on initial trials, while Phase~2 (\textit{Validation}) tests the rule on new, independent trials.

The experiments were implemented using Python and the PUREPLES framework~\cite{westh2017pureples}, which is based on NEAT-Python~\cite{McIntyre_neat-python}, and were conducted exclusively on CPUs across 14 machines with varying CPU capabilities and RAM sizes using the Optuna framework~\cite{akiba2019optuna}. This hardware variability precluded direct wall-clock time comparisons; all efficiency metrics are therefore reported in generations rather than elapsed time. All evolutionary trials were run for a maximum of 17 generations, based on prior analysis~\cite{claret2024tpe} showing only 2.7\% performance loss versus generation 20, with lower variance, justifying the per-trial computational savings for large-scale experimentation. We use MNIST as a standard benchmark for studying ES-HyperNEAT search dynamics, following prior work~\cite{claret2024tpe}. The absolute classification accuracy is not the focus of this study.

\subsection{Fitness Evaluation and Metrics}

We used the same sixteen hyperparameters as in prior work~\cite{claret2024tpe}, which provides the base search space specification. To enable broader exploration, the ranges were extended for several parameters: mutation probabilities (\texttt{conn\_add/delete\_prob}, \texttt{node\_add/delete\_prob}) from 3--4 discrete values to 0.1--0.9 in steps of 0.1; \texttt{pop\_size} from \{10, 20, 40, 60, 80, 100\} to 30--130 in steps of 10; substrate depth parameters (\texttt{initial\_depth} to 1--4, \texttt{max\_depth} to 2--6); and threshold parameters (\texttt{variance\_threshold}, \texttt{division\_threshold}) to finer granularity. This extended search space contains over 3 billion potential configurations. For the hyperparameters not included in the search space, we used the configuration file from the XOR problem experiment in the NEAT-Python library and left all other parameters at the library's default values.

The fitness of each genome was evaluated at each generation using a batch of 200 images randomly sampled from the MNIST training set, with images spread evenly across all 10 classes (20 per class). Each genome evaluation drew a fresh random batch: images were resampled independently for every genome at every generation, not fixed at trial initialization. This per-evaluation resampling prevents overfitting to any fixed subset. The class-balanced design and the use of cumulative median best fitness as the pruning metric together mitigate the resulting stochastic evaluation noise. A phenotype network's prediction was determined by the index of the maximum value in its output vector. For each correct prediction, the genome's fitness score was incremented by one, and the total score was averaged over the batch. This setup results in a baseline fitness of 0.1 for a network that makes random or constant predictions.

For the MNIST experiment, the substrate was initialized at the center of the input and output vectors. The objective value for TPE optimization was the best fitness achieved by the best-performing individual in the final generation.

While the primary pruning strategy relies on a single metric, additional metrics were recorded for detailed analysis. A key distinction is made between metrics that capture performance at a single generation versus cumulative metrics that track the historical trend of a trial. The most important metrics are defined below:

\begin{description}[leftmargin=*,labelindent=1em]
    \item[Last Generation Best Fitness:] The fitness of the single best-performing individual in the trial's final generation. This serves as the objective value for TPE optimization.

    \item[Best Fitness Over All Generations:] The peak fitness achieved by any individual in any generation throughout the trial.

    \item[Median Best Fitness:] \textbf{(Core Pruning Metric)} At generation $G$, this is the median of the set of best-per-generation fitness values recorded from generation 1 up to $G$. This cumulative statistic, chosen over alternative metrics for outlier resistance (see Section~\ref{sec:robustness}), forms the basis of our pruning rule.

    \item[Mean Best Fitness:] The mean of the set of best-per-generation fitnesses, calculated identically to the median version.
\end{description}

\subsection{Phase 1: Pruning Rule Derivation}

To determine the optimal pruning generation $G^*$ and fitness threshold $T^*$, the problem was framed as a binary classification task: predicting the final success of a trial based on its early fitness trajectory.

\subsubsection{Trial Labeling and Feature Selection}

An initial dataset was generated by sampling 30 distinct hyperparameter configurations using Optuna's random sampler, with each configuration replicated 3 times (N=90 trials). This $30 \times 3$ design balances breadth and variance: 30 unique configurations provide diverse coverage of the large search space, while three stochastic repeats per configuration reduce the risk of labeling a setting based on a single lucky/unlucky evolutionary run and yield a more stable estimate of early fitness dynamics for threshold selection. Each trial was run to completion (17 generations) and labeled based on its final median best fitness:
\begin{itemize}
    \item \textit{Positive} (Successful): Trials with a final fitness in the $80^{th}$ percentile or higher.
    \item \textit{Negative} (Unsuccessful): All other trials.
\end{itemize}

The $80^{th}$ percentile was selected to isolate the top-performing runs, in line with the Pareto principle, and to provide a stable, data-relative definition of success. In absolute terms, this cutoff corresponds to a final fitness of approximately 0.18 (about 18\% MNIST accuracy, versus the 0.1 random-guessing baseline). This choice is also justified by empirical results presented in Section \ref{sub:justification}. The predictive feature for classification was the median best fitness calculated up to a given generation.

\subsubsection{Selecting the Pruning Generation \texorpdfstring{$G^*$}{G*}}
\label{sub:prune_gen}

We identified $G^*$ as the earliest generation where successful trials consistently showed learning. For each Positive-labeled trial, we identified the first generation where median best fitness exceeded the 0.1 baseline, then computed the ceiling of the mean of these generation numbers. This method roots the pruning point in the empirical learning dynamics of high-potential trials.

\subsubsection{Selecting the Pruning Threshold \texorpdfstring{$T^*$}{T*}}

At the selected pruning generation $G^*$, a threshold sweep was conducted to find the optimal fitness cutoff $T^*$. Every unique median best fitness value observed at generation $G^*$ across the 90 trials was evaluated as a potential threshold. For each threshold, a confusion matrix was generated and the $F_1$ score was calculated. The threshold $T^*$ was selected as the one that maximized the $F_1$ score, achieving the best balance between precision and recall.

The resulting rule is simple, interpretable, and empirically grounded. Formally, let $\tilde{f}(t, g)$ denote the median of the best-per-generation fitness values for trial $t$ from generation 1 through $g$. The pruning rule is:
\begin{equation}
\label{eq:prune_rule}
\text{Prune trial } t \text{ at generation } G^* \iff \tilde{f}(t, G^*) \leq T^*
\end{equation}
where $G^*$ and $T^*$ are derived from the initial dataset as described above. In the present study, $G^* = 3$ and $T^* = 0.140$.

\subsection{Phase 2: Pruning Rule Validation and Generalization}

To validate the derived pruning rule, a new set of trials was initiated with the pruner deployed. Hyperparameter configurations were again sampled randomly using Optuna. For each trial, the median best fitness was monitored, and if it did not exceed the threshold $T^*$ at generation $G^*$, the trial was terminated.

To evaluate the pruner's accuracy and assess the rate of false negatives (i.e., promising trials that were incorrectly pruned), 90 randomly selected pruned trials were resumed and allowed to run to completion (generation 17). This counterfactual analysis produced a validation confusion matrix with precision, recall, and $F_1$ on unseen data, testing whether the pruning rule generalizes.

\section{Experimental Results}

\subsection{Justifying the Success Threshold via Sensitivity Analysis}
\label{sub:justification}

We defined a successful trial as one reaching the $80^{th}$ percentile of final fitness scores. To validate that this choice was not arbitrary, we conducted a sensitivity analysis to compare it against more permissive ($50^{th}$, $75^{th}$) and stricter ($90^{th}$) thresholds. For each of these four definitions, we derived the optimal pruning rule by finding the generation ($G^*$) and fitness threshold ($T^*$) according to our methodology. Each of these four rules was then evaluated on the independent validation set.

The results, summarized in Table~\ref{tab:percentile_sensitivity}, reveal the trade-offs between computational savings and the risk of incorrectly pruning a high-potential trial.

\begin{table}[htbp]
    \caption{
        Sensitivity analysis for the success threshold percentile. Each rule was derived on the initial dataset and evaluated on the validation set.
    }
    \label{tab:percentile_sensitivity}
    \centering
    \small
    \setlength{\tabcolsep}{4pt}
    \begin{tabular}{l c c c c c}
        \toprule
        \textbf{\makecell{Success\\Def.}} & \textbf{\makecell{Rule\\($G^*, T^*$)}} & \textbf{Recall} & \textbf{Precision} & \textbf{\makecell{$F_1$-\\Score}} & \textbf{\makecell{Savings\\(\%)}} \\
        \midrule
        $50^{th}$ & (13, 0.145) & 0.52 & 1.00 & 0.68 & 11.4 \\
        $75^{th}$ & (4, 0.143)  & 0.86 & 0.97 & 0.91 & 38.2 \\
        \textbf{$80^{th}$} & \textbf{(3, 0.140)} & \textbf{0.90} & \textbf{0.84} & \textbf{0.87} & \textbf{41.6} \\
        $90^{th}$ & (3, 0.140)  & 0.94 & 0.70 & 0.80 & 41.6 \\
        \bottomrule
    \end{tabular}
\end{table}

The $50^{th}$ percentile was judged ineffective, waiting too long to prune ($G^*=13$) and still achieving very poor recall. The $90^{th}$ percentile rule, while preserving almost all elite trials (recall = 0.94), suffered from low precision, allowing too many unsuccessful trials to pass, resulting in a lower $F_1$ score.

The primary comparison is between the $75^{th}$ and $80^{th}$ percentile rules. Although the $75^{th}$ percentile rule achieved the highest $F_1$ score (0.91), the $80^{th}$ percentile rule presented a better overall profile for our objectives. Specifically, the $80^{th}$ percentile rule:
\begin{enumerate}[leftmargin=*,labelindent=1em]
    \item \textbf{Maximized Recall:} It achieved a recall of 0.90, higher than the $75^{th}$ percentile's 0.86. For a pruning mechanism, minimizing the risk of discarding valuable ``slow starter'' trials (False Negatives) is essential, making higher recall a critical feature.
    \item \textbf{Yielded Greater Savings:} As a direct result of pruning earlier, it achieved a computational cost reduction of 41.6\%, exceeding the savings of 38.2\% from the $75^{th}$ percentile rule.
\end{enumerate}
The $80^{th}$ percentile thus gives the best trade-off between safety (high recall) and efficiency for our purposes.

\subsection{Phase 1: Derivation of the Pruning Rule}

The initial phase of the experiment aimed to answer two key questions: At which generation can we reliably distinguish between successful and unsuccessful trials? And what is the optimal fitness threshold to use for pruning at that generation?

\subsubsection{Identifying the Optimal Pruning Generation \texorpdfstring{$G^*$}{G*}}

\begin{figure}[htbp]
  \centering
  \includegraphics[width=\columnwidth]{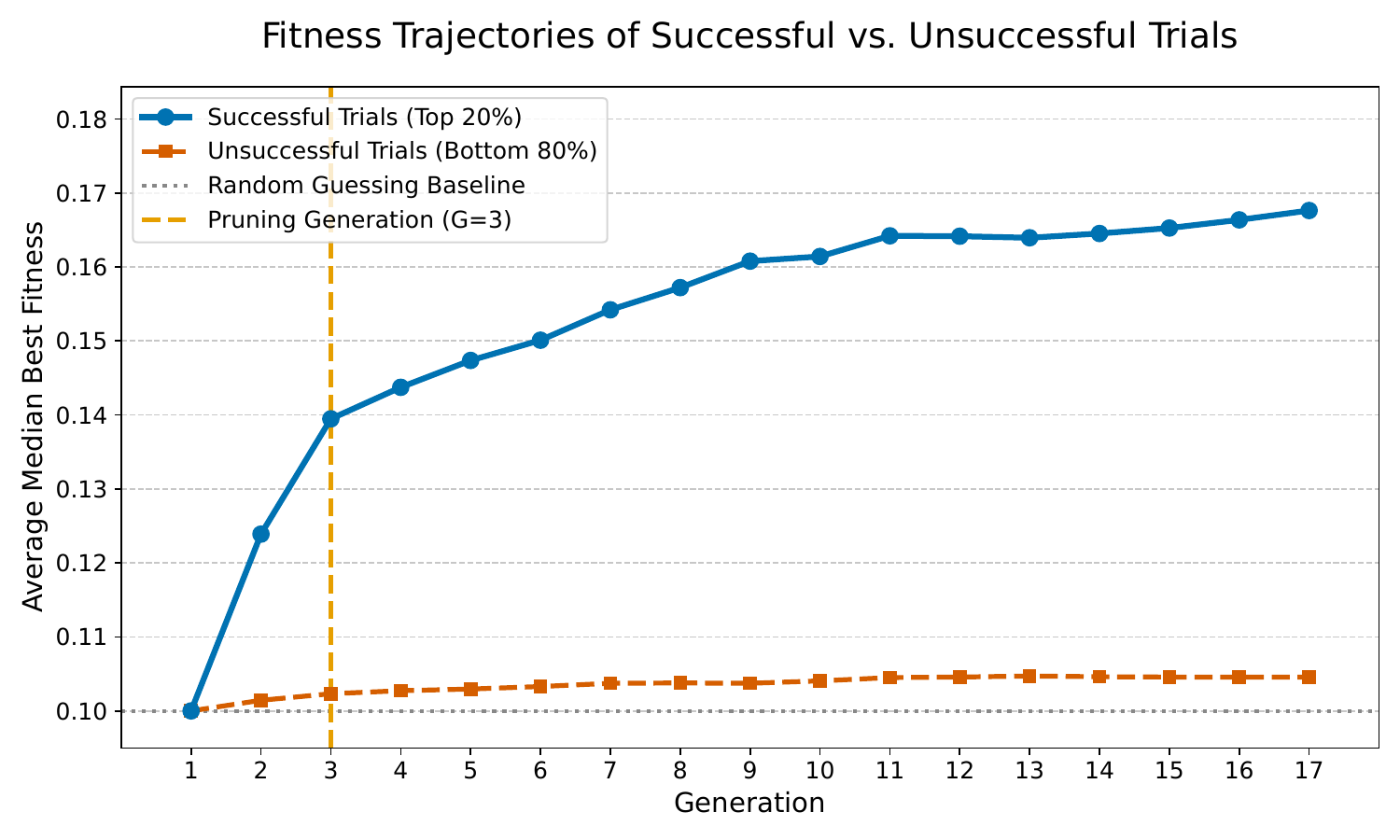}
  \caption{Fitness trajectories of ``Successful'' vs.\ ``Unsuccessful'' trials. The plot shows the mean of the median best fitness for trials labeled as ``Successful'' (n=19) and ``Unsuccessful'' (n=71).}
  \label{fig:succesful_unsuccesful}
\end{figure}

To determine the earliest point of divergence, the fitness trajectories of trials labeled ``Successful'' (top 20\%) and ``Unsuccessful'' (bottom 80\%) were analyzed. Figure~\ref{fig:succesful_unsuccesful} plots the mean of the median best fitness for both classes across the 17 generations.

As illustrated in Figure~\ref{fig:succesful_unsuccesful}, a clear separation emerges in the early generations. Following the methodology described in Section~\ref{sub:prune_gen}, we identified the generation at which successful trials reliably surpassed the 0.1 fitness baseline. This analysis yielded an optimal pruning point of $G^* = 3$. Section~\ref{sec:robustness} validates this choice statistically.

\subsubsection{Determining the Optimal Pruning Threshold \texorpdfstring{$T^*$}{T*}}

With the pruning generation fixed at $G^* = 3$, the optimal fitness threshold, $T^*$, was determined by treating the problem as a binary classification task. For every unique median fitness value observed at generation 3, which served as a candidate threshold, each trial was classified based on its early fitness. This prediction was then compared against the trial's true label.

\begin{figure}[htbp]
  \centering
  \includegraphics[width=\columnwidth]{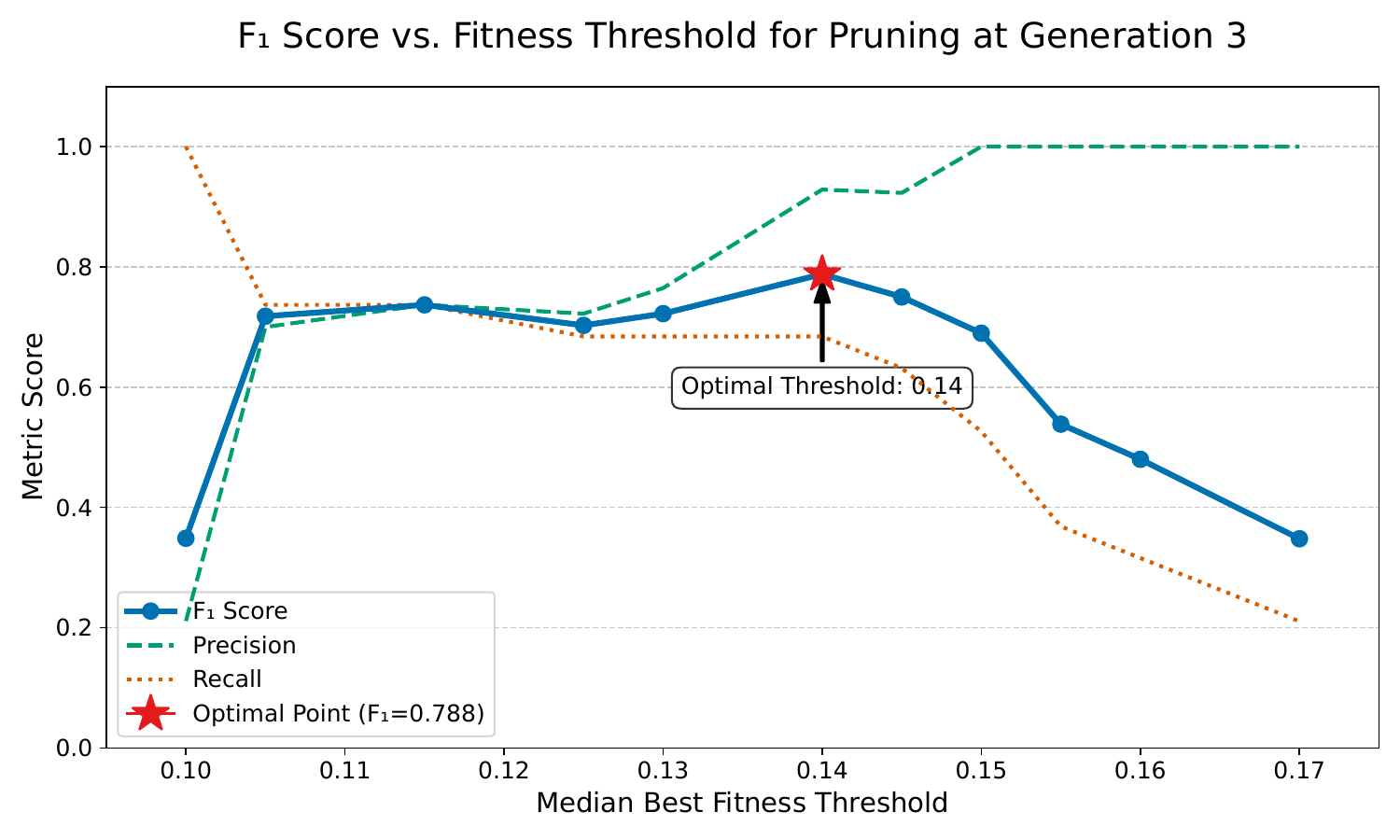}
  \caption{$F_1$ score for pruning thresholds at generation $G^*=3$. The plot shows the $F_1$ score, precision, and recall for different median best fitness thresholds. The threshold that maximizes the $F_1$ score represents the optimal trade-off between identifying unsuccessful trials and retaining successful ones.}
  \label{fig:figure_2}
\end{figure}

The analysis, shown in Figure~\ref{fig:figure_2}, identified an optimal threshold of $T^* = 0.140$. This threshold yielded a maximal $F_1$ score of $0.788$ on the derivation dataset. The $F_1 = 0.788$ reported here is computed on the derivation set (90 trials) where ``Successful'' is defined as the top 20\% final fitness. This setup both (i) yields a strongly imbalanced label distribution (19 Successful vs. 71 Unsuccessful) and (ii) introduces threshold-induced label noise near the 80th-percentile cutoff, since trials with very similar final fitness can fall on opposite sides of the boundary. The derivation $F_1$ is a conservative measure of separability at generation $G^*$, rather than an estimate under a prevalence-faithful class prior. This value distinguishes trials demonstrating genuine learning from those stuck at the 0.1 baseline. \textit{The final derived rule was therefore: Prune any trial where the median best fitness at generation 3 is less than or equal to 0.140.}

\subsection{Phase 2: Validation and Generalization}

The second phase tested the derived pruning rule on a new, independent set of trials to assess its real-world performance and generalization. The pruner was actively deployed, terminating trials that met the pruning condition.

To quantify the pruner's accuracy, 90 randomly selected pruned trials were resumed to completion to observe their true final fitness. A further 90 trials that ran to completion were selected to form a validation dataset of 180 trials. This allowed for the construction of a confusion matrix based on the pruner's decisions versus the actual outcomes, as shown in Table~\ref{tab:confusion_matrix}.

\begin{table}[h!]
    \caption{
        Confusion matrix for the pruner's performance on the validation set.
        Rows represent the actual final outcome, while columns represent the
        pruner's prediction.
    }
    \label{tab:confusion_matrix}
    \centering
    \begin{tabular}{@{} l c c @{}}
        \toprule
        & \multicolumn{2}{c}{\textbf{Predicted Outcome}} \\
        \cmidrule(l){2-3}
        \textbf{Actual Outcome} & \textbf{Successful} & \textbf{Unsuccessful (Pruned)} \\
        \midrule
        Successful              & 75 & 8  \\
        Unsuccessful            & 14 & 83 \\
        \bottomrule
    \end{tabular}
\end{table}

The matrix reveals 83 True Negatives and 75 True Positives. The primary errors were 8 False Negatives (successful trials incorrectly pruned) and 14 False Positives (unsuccessful trials that passed).

To quantitatively summarize this performance, a suite of standard metrics was calculated from the confusion matrix, as detailed in Table~\ref{tab:performance_metrics}.

\begin{table}[htbp]
    \caption{
        Key performance metrics calculated from the validation confusion matrix.
    }
    \label{tab:performance_metrics}
    \centering
    \begin{tabularx}{\columnwidth}{>{\raggedright\arraybackslash}X c}
        \toprule
        \textbf{Performance Metric} & {\textbf{Value}} \\
        \midrule
        Precision                       & 0.843 \\
        Recall (Sensitivity)            & 0.904 \\
        Specificity                     & 0.856 \\
        {$F_1$ Score}                   & 0.872 \\
        Accuracy                        & 0.878 \\
        Matthews Correl. Coeff. (MCC)   & 0.757 \\
        \bottomrule
    \end{tabularx}
\end{table}

The $F_1$ score of 0.872 and MCC of 0.757 show that the rule classifies well on unseen data. The pruner kept over 90\% of truly successful trials (recall 0.904) and correctly rejected most failing ones (specificity 0.856). Of the trials it let through, 84.3\% were successful (precision 0.843).

The validation $F_1$ (0.872) exceeds the derivation $F_1$ (0.788). This should not be read as an improvement after tuning; the pruning rule is fixed ($G^*=3$, $T^*=0.140$). Instead, the difference is expected because the validation confusion matrix is computed on a stratified evaluation set (90 pruned trials resumed + 90 trials that were not pruned and ran to completion), which changes the class prior (Table~\ref{tab:confusion_matrix}) and can make precision/recall (and therefore $F_1$) differ from the derivation setting.

These validation results confirm that the empirically derived pruning rule generalizes to unseen trials from the same search space.

\subsection{Computational Efficiency}

Beyond classification accuracy, the pruning mechanism aims to reduce computational cost. We quantified this by comparing total generations evaluated in our validation set against the number required without pruning.

A trial terminated by the pruner runs for only 3 generations, whereas completion requires 17 generations. Based on Table~\ref{tab:confusion_matrix}, pruning reduced total cost from 3,060 generations (180 trials $\times$ 17 gens) to 1,786 generations ((91 pruned $\times$ 3) + (89 completed $\times$ 17)), achieving a 41.6\% reduction while retaining over 90\% of successful trials.

\textbf{Comparison with Hyperband.} To assess the performance of our custom pruner, we conducted a parallel experiment using a standard pruner from the Optuna framework. We ran a TPE search combined with the HyperbandPruner (HbP), configured with a \texttt{min\_resource=2}, \texttt{max\_resource=17}, and a \texttt{reduction\_factor=2}. This configuration creates four successive ``rungs'' where trials are evaluated and potentially pruned.

The performance of our data-driven pruner was compared with this Hyperband baseline across several key metrics, summarized in Table~\ref{tab:hyperband_comparison}.

\begin{table}[htbp]
    \caption{
        Comparative performance between our custom pruner, the standard Hyperband pruner, and the unpruned TPE baseline.
    }
    \label{tab:hyperband_comparison}
    \centering
    \footnotesize
    \setlength{\tabcolsep}{3pt}
    \begin{tabular}{l c c c}
        \toprule
        \textbf{Performance Metric} & \makecell{Our \\ Method} & \makecell{Hyper- \\ band} & Baseline \\
        \midrule
Best Final Fitness Found & 0.225 & 0.250 & 0.275 \\
Avg. Top 5\% Fitness & 0.206 & 0.195 & 0.241 \\
Total Cost (Gens) & 10,656 & 29,574 & 40,260 \\
Trials to 0.25 & Not Reached & 569 & 359 \\
\bottomrule
    \end{tabular}
\end{table}

Peak fitness and efficiency trade off against each other. Hyperband found a slightly higher peak fitness (0.250 vs. 0.225), but our pruner produced better configurations on average: the top 5\% of our trials averaged 0.206 vs. Hyperband's 0.195. Hyperband's broader search occasionally hits a higher maximum, but our method concentrates resources on the productive part of the search space.

The cost difference is large: 10,656 total generations for our pruner vs. 29,574 for Hyperband, a 64\% reduction. This efficiency gain comes at the expense of the lower peak fitness noted above, but the domain-specific rule concentrates budget on high-potential trials more effectively than Hyperband's fixed schedule.

Our method trades peak fitness for search efficiency in this problem domain. We did not match the baseline's peak fitness, which warrants further investigation into how the pruner behaves with larger trial budgets.

\textbf{Statistical Comparison.} To provide a rigorous quantitative comparison of the overall effectiveness of our pruner, the standard Hyperband pruner, and the unpruned TPE baseline, we analyzed the distributions of all final fitness scores from every completed trial in each study. A one-way Analysis of Variance (ANOVA) was conducted (using all completed trials from the current study combined with the 2,013 trials from prior work~\cite{claret2024tpe}), which revealed a statistically significant difference in the mean final fitness achieved across the three methods (\textit{F}(2, 3676) = 1090.7, $p < .001$). At $N = 3{,}679$, the Central Limit Theorem ensures robustness of the ANOVA to non-normality~\cite{kwak2017central}. A confirmatory Kruskal--Wallis test on the surviving archived data subset ($N = 2{,}695$) confirmed this result ($H(2) = 421.5$, $p < .001$, $\varepsilon^2 = 0.156$), verifying robustness to distributional assumptions without requiring normality. The agreement between parametric (ANOVA) and non-parametric (Kruskal--Wallis) omnibus tests validates the subsequent Tukey HSD post-hoc comparisons. For the paired $t$-tests (Table~\ref{tab:metric_comparison}), $N=90$ per group similarly ensures CLT robustness~\cite{kwak2017central}.

Tukey's HSD post-hoc test (Table~\ref{tab:tukey_results}) shows that our pruner significantly outperforms Hyperband in mean population fitness, with a large effect size (Cohen's \textit{d} = 1.884).

No significant difference appeared between our pruner and the unpruned TPE baseline (\textit{p} = 0.069). The pruner finds equivalent solutions at 41.6\% lower cost.

\begin{table}[h!]
    \caption{
        Tukey's HSD post-hoc test results comparing the three methods.
    }
    \label{tab:tukey_results}
    \centering
    \footnotesize
    \setlength{\tabcolsep}{3pt}
    \begin{tabular}{lcccc}
        \toprule
        \textbf{Comparison} & \textbf{Mean Diff} & \textbf{\textit{p}-value} & \textbf{Sig.} & \textbf{Cohen's \textit{d}} \\
        \midrule
        Our vs. HbP & 0.052  & $< .001$ & Yes & 1.884 \\
        Baseline vs. HbP     & 0.058  & $< .001$ & Yes & 1.555 \\
        Our vs. Baseline    & -0.006 & 0.069    & No  & -     \\
        \bottomrule
    \end{tabular}
\end{table}

\textbf{Equivalence Testing.} While the ANOVA showed no significant difference, this is not strong evidence of equivalence. To make a more rigorous claim, we performed a Two One-Sided Tests (TOST) for equivalence. We defined a practical equivalence margin ($\Delta$) of $\pm$2\% absolute fitness (0.02). Given that peak fitness in our experiments is approximately 0.275, this margin corresponds to roughly four additional correctly classified images per 200-image batch, a practically insignificant difference.

Table~\ref{tab:tost_results} shows the TOST result is significant ($p < .001$), formally rejecting non-equivalence. The 41.6\% cost reduction does not degrade solution quality.

\begin{table}[h!]
    \caption{
        Results of the TOST equivalence test comparing our method to the unpruned baseline.
    }
    \label{tab:tost_results}
    \centering
    \small
    \begin{tabular}{lc}
        \toprule
        \textbf{Equivalence Test Metric} & \textbf{Value} \\
        \midrule
        Equivalence Margin ($\Delta$) & $\pm$0.020 \\
        Observed Mean Difference & -0.0059 \\
        TOST \textit{p}-value & $< .001$ \\
        \textbf{Conclusion} & \textbf{Statistically Equivalent} \\
        \bottomrule
    \end{tabular}
\end{table}

\subsection{Analysis of Pruning Errors}

\begin{figure}[htbp]
  \centering
  \includegraphics[width=\columnwidth]{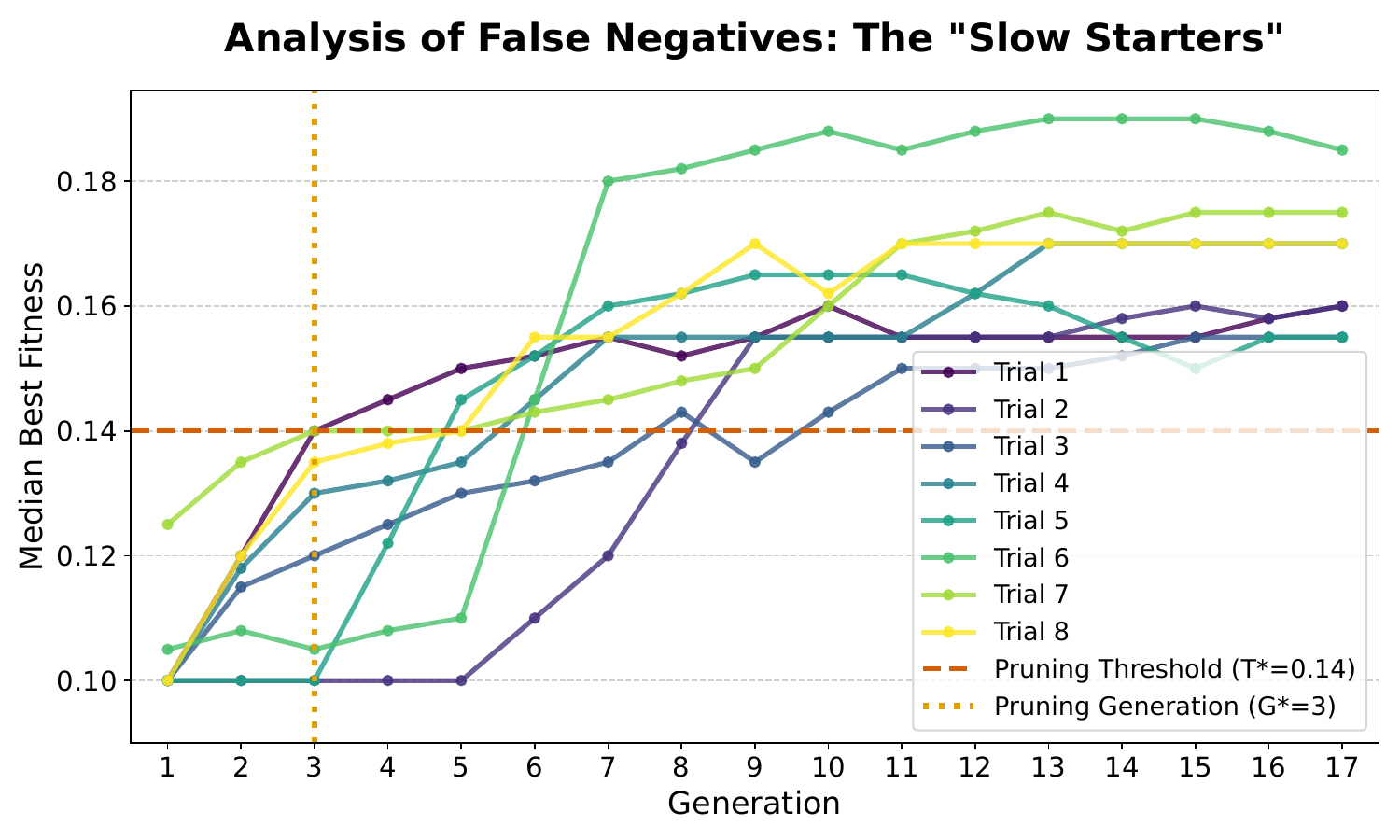}
  \caption{
      Fitness trajectories of the 8 successful trials that were incorrectly pruned (False Negatives). These trials failed to surpass the fitness threshold of 0.140 (red dashed line) by the pruning generation $G^*{=}3$ (vertical line).
      }
  \label{fig:false_negatives}
\end{figure}

Eight successful trials were incorrectly killed (False Negatives). Figure~\ref{fig:false_negatives} plots their full fitness trajectories.

These trials are ``slow starters'': they eventually reach high fitness, but their median best fitness at generation 3 sat below the 0.140 threshold. An aggressive early cutoff inevitably loses a few of these late-blooming runs. The rule accepts this risk because the 41.6\% cost savings outweigh the loss of 8 out of 83 successful trials.

\textbf{False Positives.} The 14 false positives (trials that passed early but ultimately failed) result from batch-sampling variance and the relative success definition. These errors reduce computational savings rather than solution quality, as reflected in the precision of 0.843.

\subsection{Robustness Validation}
\label{sec:robustness}

\textbf{Metric Selection.} Because the $80^{th}$ percentile of \textit{last best fitness} remained at baseline 0.1 (random guessing), preventing meaningful differentiation, we adopted \textit{median best fitness}. Paired $t$-tests confirmed no significant difference between metrics ($p > 0.05$, Cohen's $d < 0.2$; Table~\ref{tab:metric_comparison}). At $N = 90$ per group, the $t$-test is robust to non-normality by the Central Limit Theorem~\cite{kwak2017central}.

\begin{table}[htbp]
    \caption{
        Results of paired-samples $t$-tests comparing final performance metrics
        (\textit{Last Median Best} vs. \textit{Last Best}) within each experimental group.
    }
    \label{tab:metric_comparison}
    \centering
    \setlength{\tabcolsep}{4pt}
    \begin{tabular}{
        l
        S[table-format=-1.2]
        S[table-format=0.3]
        S[table-format=-1.2]
    }
        \toprule
        \textbf{Experimental Group} &
        \textbf{$t$-statistic} &
        \textbf{$p$-value} &
        \multicolumn{1}{c}{\textbf{Cohen's $d$}} \\
        \midrule
        Random Search (N=90)        & -0.88 & 0.378 & -0.13 \\
        Random Unpruned (N=90)      & -0.65 & 0.519 & -0.10 \\
        Random Pruned (N=90)        & -1.00 & 0.321 & -0.15 \\
        \bottomrule
    \end{tabular}
\end{table}

\textbf{Validation of $G^*=3$.} To ensure the selected pruning generation of $G^*=3$ was truly optimal and not merely an artifact of the dataset, we statistically validated its performance against adjacent generations. For each candidate generation ($G \in \{2, 3, 4\}$), we used the pruning threshold derived from our initial, large-scale analysis ($T^*=0.122$ for $G=2$, $T^*=0.140$ for $G=3$, and $T^*=0.143$ for $G=4$).

We then evaluated each of these three fixed rules on our independent, 180-trial validation set. To assess statistical significance, we applied two complementary tests: McNemar's exact test for paired classifier comparison and a paired bootstrap test (1,000 resamples) for the $F_1$-score difference.

Table~\ref{tab:gen_robustness} shows that $G^*=3$ is significantly better than $G^*=2$. McNemar's test confirms a significant difference ($p = 0.022$), with $G^*=3$ correctly classifying 11 trials that $G^*=2$ misclassified, while only 2 trials showed the reverse. The paired bootstrap test agrees: the 95\% CI for the $F_1$ difference ($G^*=3 - G^*=2$) is [0.012--0.098], excluding zero ($p = 0.003$). Waiting one extra generation produces a more reliable decision.

$G^*=3$ and $G^*=4$ do not differ significantly (McNemar's $p = 0.375$), but $G^*=3$ intervenes one generation sooner. It is the earliest generation that achieves top classification performance.

\begin{table}[htbp]
    \caption{
        $F_1$-scores and 95\% confidence intervals for the paper's pruning rules, evaluated on the curated 180-trial validation set. Confidence intervals were calculated via bootstrapping (N=1,000 resamples).
    }
    \label{tab:gen_robustness}
    \centering
    \begin{tabular}{ccc}
        \toprule
        \textbf{Pruning Gen. ($G^*$)} & \textbf{$F_1$-Score} & \textbf{95\% Confidence Interval} \\
        \midrule
        2 & 0.819 & [0.754--0.878] \\
        \textbf{3} & \textbf{0.872} & \textbf{[0.817--0.924]} \\
        4 & 0.869 & [0.812--0.917] \\
        \midrule
        \multicolumn{3}{l}{\textit{Paired difference ($G^*=3 - G^*=2$): 0.053 [0.012--0.098]}} \\
        \bottomrule
    \end{tabular}
\end{table}

\subsection{Generalizability Test: Simulation on a Converged Search Space}

To test boundary conditions, we simulated our rule on the 824 late-stage trials from prior work~\cite{claret2024tpe} that contained per-generation data. This tests whether an early-stage pruner remains effective on a converged, pre-filtered search space.

\begin{figure}[htbp]
  \centering
  \includegraphics[width=\columnwidth]{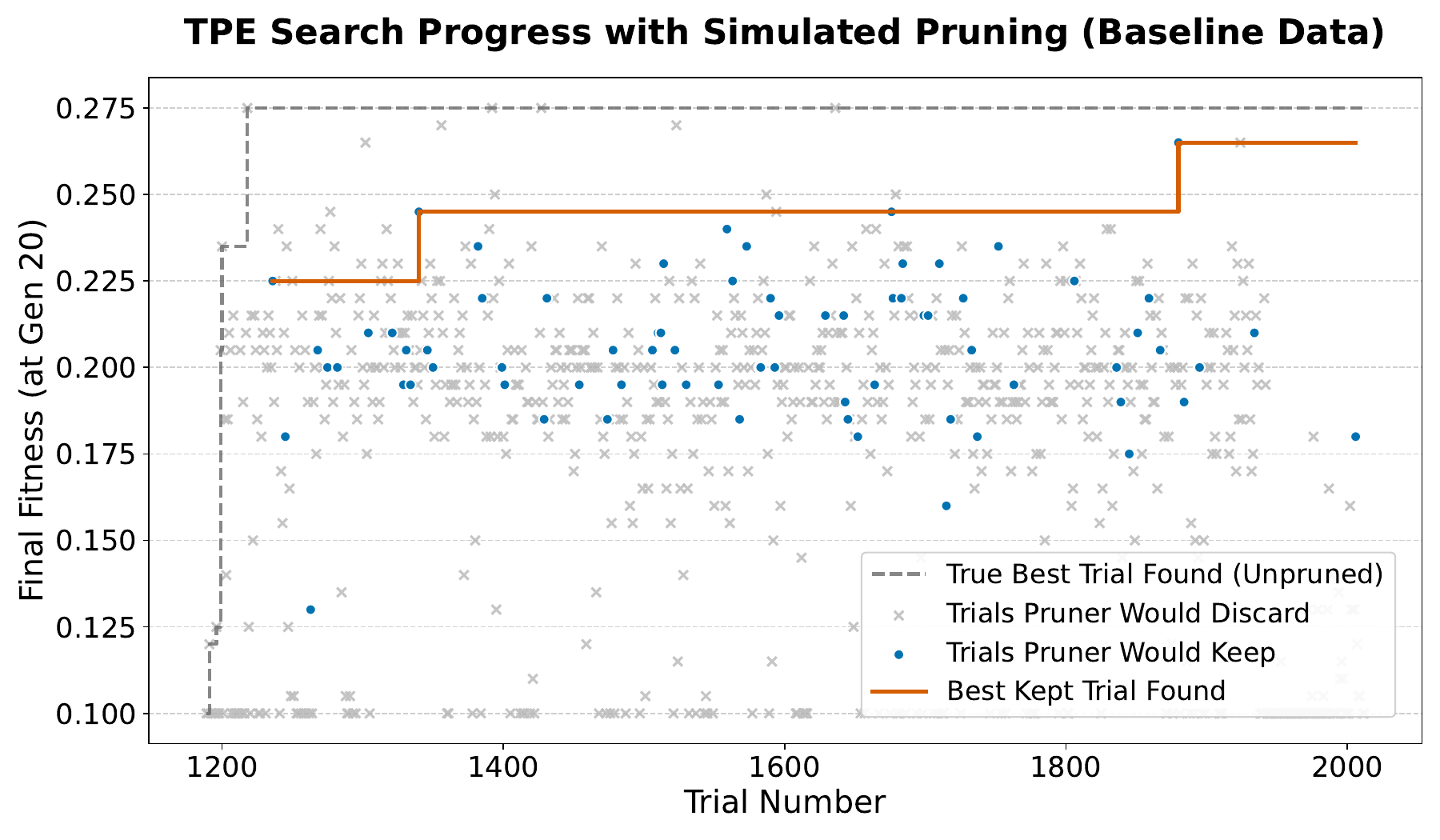}
  \caption{
      TPE search progress with our pruner simulated on the prior dataset. Gray crosses mark trials the pruner would have dismissed; blue dots mark trials it would have kept. The dashed gray line shows the true best fitness found over all trials (unpruned baseline), while the solid orange line shows the best fitness achieved using only kept trials.
      }
  \label{fig:baseline_tpe_progress}
\end{figure}

Table~\ref{tab:baseline_simulation} summarizes the outcome. The pruner still separates fit from unfit trials (\textit{t}(822)=15.9, $p < .001$), with 73\% simulated savings. But the aggressive threshold is too strict for this elite population: recall dropped to 31.1\%, and the single best trial would have been killed, reducing peak fitness from 0.275 to 0.245 (Figure~\ref{fig:baseline_tpe_progress}).

\begin{figure}[htbp]
  \centering
  \includegraphics[width=\columnwidth]{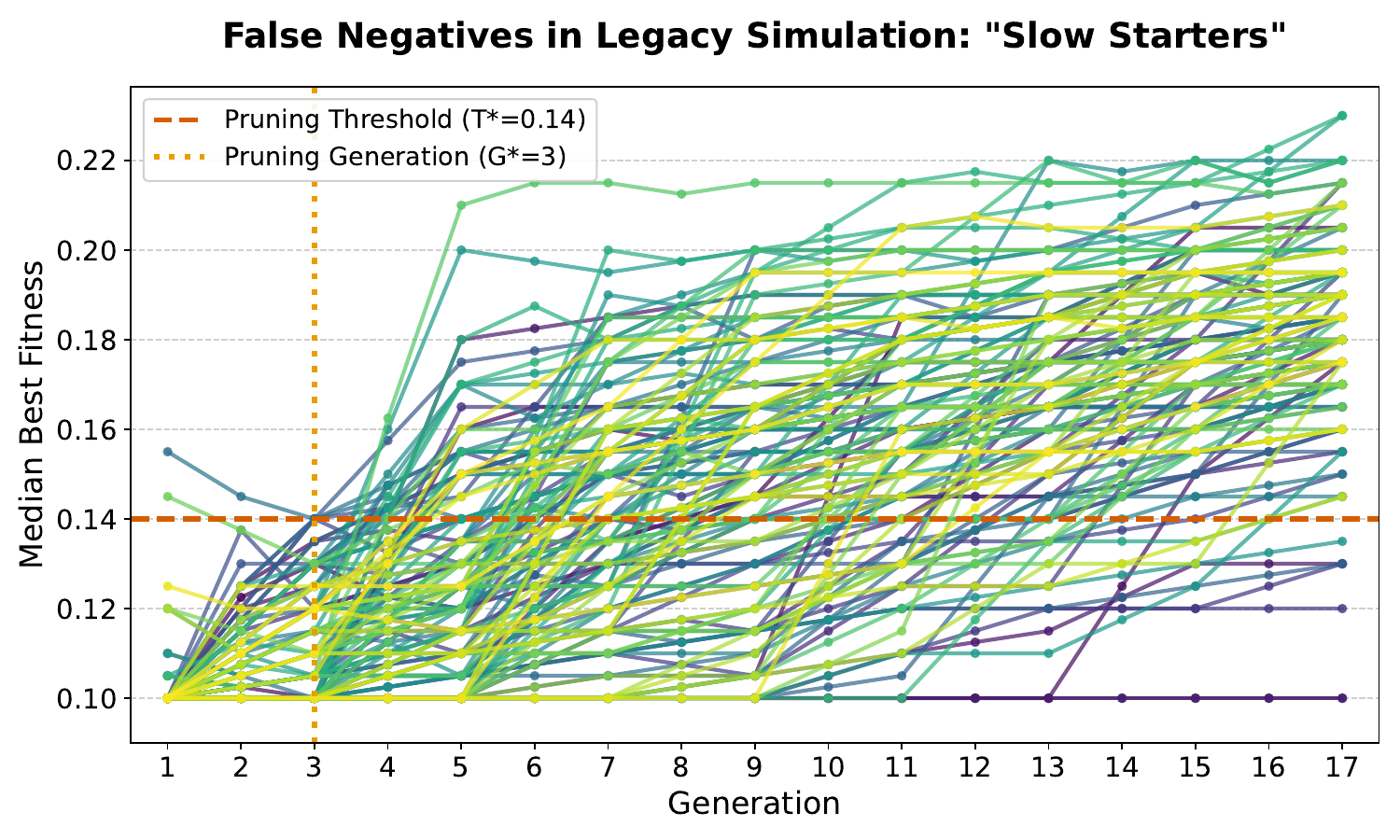}
  \caption{
      Fitness trajectories of the ``slow starter'' trials incorrectly pruned during the baseline simulation (False Negatives).
      }
  \label{fig:baseline_slow_starters}
\end{figure}

Figure~\ref{fig:baseline_slow_starters} shows the fitness trajectories of these false negatives, the ``slow starters'' characteristic of converged search spaces. The rule is specialized for random initial populations, motivating adaptive pruners that adjust strictness as search converges.

\begin{table}[h!]
    \caption{
        Results of a counterfactual simulation applying our $G^*{=}3$ pruning rule to the final 824 trials of a prior dataset.
    }
    \label{tab:baseline_simulation}
    \centering
    \small
    \begin{tabular}{lc}
        \toprule
        \textbf{Late-Stage Simulation Metric} & \textbf{Value} \\
        \midrule
        Simulated Computational Savings & 73.2 \% \\
        $F_1$-Score on Late-Stage Trials & 0.402 \\
        Recall on Top 20\% Late-Stage Trials & 31.1 \% \\
        Mean Fitness of Kept Trials & 0.1870 \\
        Mean Fitness of Pruned Trials & 0.1500 \\
        Statistical Separation (\textit{p}-value) & $< .001$ \\
        \bottomrule
    \end{tabular}
\end{table}

\section{Discussion}
\label{sec:discussion}

The pruning rule, derived from the fitness dynamics of a random initial population, achieved $F_1 = 0.872$ and 41.6\% computational savings on the validation set. Against Hyperband, it cut cost by 64.0\% while producing higher mean fitness (Cohen's \textit{d} = 1.884), though Hyperband found a slightly higher peak. Domain-specific pruning outperforms general-purpose schedules when the algorithm's failure mode is predictable from early trajectories.

Schedule-based pruners (SHA, Hyperband) apply fixed elimination ratios at predetermined resource budgets regardless of domain. Our approach instead derives its stopping criterion from the fitness dynamics of the specific algorithm-task pair. $T^*$ and $G^*$ encode a concrete fact about ES-HyperNEAT on MNIST: trials either show learning by generation 3 or never do, and no generic schedule can capture this. The same domain specificity that yields the 64\% efficiency advantage over Hyperband also explains the limitation on converged populations: the rule is calibrated for initial random exploration, not late-stage refinement.

The binary learn-or-don't behavior that makes early stopping possible follows from ES-HyperNEAT's indirect encoding. The CPPN must produce weight and connectivity patterns that route information from input nodes through the quadtree-generated substrate to output nodes. At generation~0, 93\% of our 180 validation trials sit at the random-guessing baseline of 0.1, and 29\% never leave it across all 17 generations. Trials still at baseline at generation~3 remain stuck for the full run 87\% of the time: small NEAT mutations (adding a node, toggling a connection) rarely bridge the gap between a substrate with no working signal path and one that propagates input information to outputs. Successful trials show an immediate jump at generation~1 (mean fitness rises from 0.10 to 0.14) and then plateau: mean fitness changes by less than 0.005 between generation~1 and generation~16. The CPPN-to-substrate mapping produces a fitness surface with a large flat region at random guessing and a separate region of viable configurations, and whether a trial lands in one or the other is largely determined at initialization.

\subsection{Limitations and Future Directions}

The counterfactual simulation on late-stage trials (recall 31.1\%) shows that a rule calibrated for diverse populations is too strict once the search converges and ``slow starters'' become more common, even though statistical separability holds ($p < .001$).

Adaptive pruning that starts aggressive and relaxes as the population improves would address this~\cite{MohrVanRijn2021LCCV}. Possible directions include median-based rules that compare intermediate results against completed trials, learning-curve forecasting~\cite{domhan2015speeding,Swersky2014FreezeThaw}, and hybrid methods like BOHB and DEHB that reallocate budget to late bloomers~\cite{Falkner2018BOHB,Awad2021DEHB}.

The optimal pruning parameters likely depend on task-specific dynamics, so transferability is uncertain. Future work should test on other benchmarks and explore metalearning frameworks that derive problem-specific stopping rules.

The mechanistic account in Section~\ref{sec:discussion} rests on aggregate trajectory statistics rather than direct landscape measurement. Characterizing the ES-HyperNEAT fitness surface more directly, for example by measuring the basin of attraction at the baseline plateau or the genetic distance between stuck and escaped CPPN initializations, would strengthen the landscape claim.

\section{Conclusion}

We derived a data-driven early-stopping rule for ES-HyperNEAT that achieves $F_1 = 0.872$ and cuts computational cost by 41.6\% without degrading fitness quality. Simulation on converged search spaces exposed the boundary conditions of static rules and points toward adaptive mechanisms that relax the threshold as the population improves. The specific thresholds ($G^* = 3$, $T^* = 0.140$) are tied to ES-HyperNEAT on MNIST, but the methodology (framing pruning as classification on early fitness dynamics) applies to any evolutionary algorithm whose failures are diagnosable from early trajectory statistics.

\ifanonymous
% Acknowledgment hidden for anonymous submission
\else
\section*{Acknowledgment}
Use of Generative AI: In accordance with IEEE policy, we disclose that Claude~\cite{claude2026} (Anthropic) was used to assist with creating figure visualizations and revising drafts to improve grammar, spelling, and consistency of writing style. All AI-generated content was reviewed, verified, and edited by the authors, who take full responsibility for the final work.
\fi


\begin{thebibliography}{00}

\bibitem{stanley2002evolving} K. O. Stanley and R. Miikkulainen, ``Evolving neural networks through augmenting topologies,'' \emph{Evolutionary Computation}, vol. 10, no. 2, pp. 99--127, 2002.

\bibitem{risi2012enhanced} S. Risi and K. O. Stanley, ``An enhanced hypercube-based encoding for evolving the placement, density, and connectivity of neurons,'' \emph{Artificial Life}, vol. 18, no. 4, pp. 331--363, 2012.

\bibitem{bergstra2011algorithms} J. Bergstra, R. Bardenet, Y. Bengio, and B. K\'egl, ``Algorithms for hyper-parameter optimization,'' in \emph{Advances in Neural Information Processing Systems}, vol. 24, 2011.

\bibitem{claret2024tpe} R. Claret, M. O'Neill, P. Cotofrei, and K. Stoffel, ``Investigating hyperparameter optimization and transferability for {ES-HyperNEAT}: A {TPE} approach,'' in \emph{Proc. Genetic and Evolutionary Computation Conference Companion (GECCO '24 Companion)}. New York, NY, USA: ACM, 2024, pp. 1879--1887.

\bibitem{shahriari2015taking} B. Shahriari, K. Swersky, Z. Wang, R. P. Adams, and N. de Freitas, ``Taking the human out of the loop: A review of {Bayesian} optimization,'' \emph{Proc. IEEE}, vol. 104, no. 1, pp. 148--175, 2016.

\bibitem{bergstra2012random} J. Bergstra and Y. Bengio, ``Random search for hyper-parameter optimization,'' \emph{J. Mach. Learn. Res.}, vol. 13, no. 10, pp. 281--305, 2012.

\bibitem{jamieson2016non} K. Jamieson and A. Talwalkar, ``Non-stochastic best arm identification and hyperparameter optimization,'' in \emph{Artificial Intelligence and Statistics}. PMLR, 2016, pp. 240--248.

\bibitem{li2017hyperband} L. Li, K. Jamieson, G. DeSalvo, A. Rostamizadeh, and A. Talwalkar, ``Hyperband: A novel bandit-based approach to hyperparameter optimization,'' \emph{J. Mach. Learn. Res.}, vol. 18, no. 185, pp. 1--52, 2018.

\bibitem{domhan2015speeding} T. Domhan, J. T. Springenberg, and F. Hutter, ``Speeding up automatic hyperparameter optimization of deep neural networks by extrapolation of learning curves,'' in \emph{IJCAI}, vol. 15, 2015, pp. 3460--3468.

\bibitem{akiba2019optuna} T. Akiba, S. Sano, T. Yanase, T. Ohta, and M. Koyama, ``Optuna: A next-generation hyperparameter optimization framework,'' in \emph{Proc. 25th ACM SIGKDD Int. Conf. Knowledge Discovery \& Data Mining}, 2019, pp. 2623--2631.

\bibitem{stanley2007compositional} K. O. Stanley, ``Compositional pattern producing networks: A novel abstraction of development,'' \emph{Genetic Programming and Evolvable Machines}, vol. 8, no. 2, pp. 131--162, 2007.

\bibitem{westh2017pureples} A. Westh and S. K. Munck, ``Pureples -- Pure Python library for {ES-HyperNEAT},'' 2017. [Online]. Available: \url{https://github.com/ukuleleplayer/pureples}

\bibitem{McIntyre_neat-python} A. McIntyre, M. Kallada, C. G. Miguel, C. Feher de Silva, and M. L. Netto, ``{neat-python}, version 2.1.0,'' doi:10.5281/zenodo.19024753. [Online]. Available: \url{https://github.com/CodeReclaimers/neat-python}

\bibitem{kwak2017central} S. G. Kwak and J. H. Kim, ``Central limit theorem: The cornerstone of modern statistics,'' \emph{Korean J. Anesthesiol.}, vol. 70, no. 2, pp. 144--156, 2017.

\bibitem{MohrVanRijn2021LCCV} F. Mohr and J. N. van Rijn, ``Fast and informative model selection using learning curve cross-validation,'' \emph{IEEE Trans. Pattern Anal. Mach. Intell.}, vol. 45, no. 8, pp. 9669--9680, 2023.

\bibitem{Swersky2014FreezeThaw} K. Swersky, J. Snoek, and R. P. Adams, ``Freeze-thaw {Bayesian} optimization,'' 2014, arXiv:1406.3896.

\bibitem{Falkner2018BOHB} S. Falkner, A. Klein, and F. Hutter, ``{BOHB}: Robust and efficient hyperparameter optimization at scale,'' in \emph{Int. Conf. Machine Learning}. PMLR, 2018, pp. 1437--1446.

\bibitem{Awad2021DEHB} N. Awad, N. Mallik, and F. Hutter, ``{DEHB}: Evolutionary hyperband for scalable, robust and efficient hyperparameter optimization,'' in \emph{Proc. 30th Int. Joint Conf. Artificial Intelligence (IJCAI-21)}, 2021, pp. 2147--2153.

\bibitem{claude2026} Anthropic, ``Claude,'' 2026. [Online]. Available: \url{https://www.anthropic.com/claude}

\end{thebibliography}
\end{document}